\documentclass[letterpaper, 10 pt, conference]{ieeeconf}  % Comment this line out if you need a4paper

\IEEEoverridecommandlockouts                              % This command is only needed if 
\usepackage{amsmath} % assumes amsmath package installed
\usepackage{amssymb}  % assumes amsmath package installed
\usepackage{cite}
\usepackage{graphicx}
\usepackage{siunitx}
\usepackage{url}

\title{\LARGE \bf
Design and Control of a Ballbot Drivetrain with High Agility, Minimal Footprint, and High Payload}

\author{Chenzhang Xiao$^{1}$, 
        Mahshid Mansouri$^{1}$, 
        David Lam$^{1,2}$, 
        Joao Ramos$^{1}$,~\IEEEmembership{Member,~IEEE} \\
        and Elizabeth T. Hsiao-Wecksler$^{1}$, ~\IEEEmembership{Member,~IEEE} % <-this % stops a space
\thanks{* Research supported by National Science Foundation National Robotics Initiative (award number 2024905).}% <-this % stops a space
\thanks{$^{1}$C. Xiao, M. Mansouri, D. Lam, J. Ramos, and E.T. Hsiao-Weckler are with the Department of Mechanical Science and Engineering, University of Illinois at Urbana-Champaign, IL 61820, USA (ethw@illinois.edu}%
\thanks{$^{2}$ D. Lam is also with the Department of Mechanical Engineering at the University of Michigan, Ann Arber, MI 48109, USA. }%
}

\begin{document}

\maketitle
\thispagestyle{empty}
\pagestyle{empty}

\begin{abstract}

This paper presents the design and control of a ballbot drivetrain that aims to achieve high agility, minimal footprint, and high payload capacity while maintaining dynamic stability. Two hardware platforms and analytical models were developed to test design and control methodologies. The full-scale ballbot prototype (MiaPURE) was constructed using off-the-shelf components and designed to have agility, footprint, and balance similar to that of a walking human. The planar inverted pendulum testbed (PIPTB) was developed as a reduced-order testbed for quick validation of system performance.
We then proposed a simple yet robust LQR-PI controller to balance and maneuver the ballbot drivetrain with a heavy payload. This is crucial because the drivetrain is often subject to high stiction due to elastomeric components in the torque transmission system. This controller was first tested in the PIPTB to compare with traditional LQR and cascaded PI-PD controllers, and then implemented in the ballbot drivetrain. The MiaPURE drivetrain was able to carry a payload of 60 kg, achieve a maximum speed of 2.3 m/s, and come to a stop from a speed of 1.4 m/s in 2 seconds in a selected translation direction.
Finally, we demonstrated the omnidirectional movement of the ballbot drivetrain in an indoor environment as a payload-carrying robot and a human-riding mobility device. Our experiments demonstrated the feasibility of using the ballbot drivetrain as a universal mobility platform with agile movements, minimal footprint, and high payload capacity using our proposed design and control methodologies.

%Analytical models and physical testbeds were developed to help with the design selection of mechanical components and the development of a robust controller.
\end{abstract}


\section{Introduction} \label{study0:intro}

\begin{figure}
\begin{center}
    \includegraphics[scale = 0.5] 
  {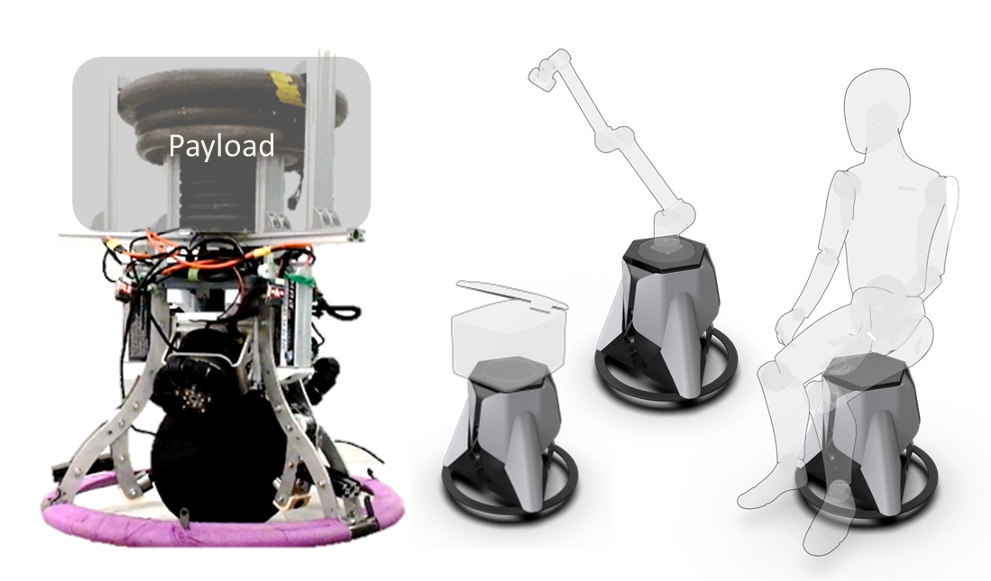}
    \caption{ Prototype of MiaPURE drivetrain balancing with a 60 kg payload, and CAD renderings of the proposed mobility platform with potential top modules for package delivery, mobile manipulation, and human riding.}
    \label{fig0:vision}
\end{center}
\end{figure}

In this study, we proposed the development of a modular ballbot drivetrain as a universal mobility platform (Fig. \ref{fig0:vision}). Ballbots, or ball balancing robots, are a family of dynamically stable mobile robots riding on top of a ball, or a spherical wheel \cite{lauwers2005one,kumagai2008development,leutenegger2010modeling}. The unique drivetrain design has a nonholonomic constraint\cite{lauwers2006dynamically}: it enables omnidirectional maneuverability such that the device can move, or translate, in any direction and spin around its vertical axis independently.

Our goal was to create a ballot drivetrain that can be configured with different top modules and controlled through a remote control device or physical human-robot interaction. The potential applications for this platform could range from standalone tasks, such as package delivery and surveillance, to collaborative tasks such as mobile manipulation and human riding (Fig. \ref{fig0:vision}). Among them, the human riding task presents unique challenges due to the high load capacity, agile locomotion, and safety requirements involved. To address these challenges, we explored and validated a drivetrain prototype of a mobility platform called Modular interactive adaptive Personal Unique Rolling Experience (MiaPURE). MiaPURE is a ballbot with a minimal footprint that can carry the weight of a human, navigate in a constrained space with its omnidirectional maneuverability, and enable intuitive control via physical interactions, such as maneuvering via torso leaning for human riding tasks.

% \subsection{Challenges in the mechanical system}
Building a ballbot with high agility, minimal footprint, and high load capacity is a nontrivial task due to a lack of standard guidelines for selecting or customizing drivetrain components, including actuators, omniwheels, and the spherical wheel. A few previous ballbots with load capacity similar to the human weight have been developed. These devices include the CMU ballbot \cite{lauwers2006dynamically}, OmniRide \cite{hoshino2013omniride}, OmniRide2 \cite{hoshino2016design}, and Ball Segway \cite{ ba2016balancing}. However, few benchmark results have been provided for these ballbot drivetrains based on performance specifications, such as the maximum speed and minimum braking time. Moreover, most of these ballbots lacked the compactness required for navigation in a constrained indoor environment. Therefore, we needed to re-visit design and control methodologies that could address these challenges and serve as a benchmark for future research.

%\subsection{Challenges in the control system}
Controlling a ballbot for balancing and maneuvering is also challenging; not only due to its nonlinear, unstable zero dynamics, but also the unmodeled friction and dynamics in the torque transmission system. Model-based optimal controllers (such as the linear quadratic regulator, or LQR)\cite{leutenegger2010modeling} often fail to handle the unmodeled stiction in the system. On the other hand, empirical controllers such as the cascaded proportional-integral-proportional-derivative (PI-PD) controller \cite{lauwers2006dynamically,kumagai2008development, nagarajan2009state} lack optimality and are often hard to tune and less robust to changes to the system parameter such as the payload weight. In this case, a model-based controller with stiction compensation capability would be desirable, especially for the ballbot drivetrain with heavy payloads. A few researchers implemented a sliding mode controller to handle uncertainties and unmodeled dynamics \cite{do2020robust, pham2019combined}, but it is often subject to chattering and difficulty in tuning when implemented in physical hardware \cite{young1999control}.

%\subsection{Overview}
The main contribution of this study is the mechanical design of a minimal footprint, high payload ballbot drivetrain, and investigation into a cascaded LQR-PI controller to achieve high agility.
The mechanical design of the MiaPURE drivetrain and a reduced-order planar inverted pendulum testbed (PIPTB) are detailed in Section \ref{study0:design}. Section \ref{study0:model} reviews the system modeling and trajectory optimization for a safety-critical braking task for later benchmark experiments. Section \ref{study0:control} presents the cascaded LQR-PI controller, which combines the advantages of both model-based LQR and PI controller to compensate for unmodeled friction in the system. The PIPTB is used to preliminarily validate the controller performance. Section \ref{study0:exp} evaluates the performance of the full-sized MiaPURE drivetrain with a heavy payload, including the maximum speed and minimum braking time in multiple translation directions, along with demonstrations of payload carrying and human riding. Section \ref{study0:discussion} discusses insights from the experiments, system limitations, typical failure modes, and future work directions, followed by a conclusion in Section \ref{study0:conclusion}.

\section{HARDWARE PLATFORMS}\label{study0:design}

Two hardware platforms were developed: the MiaPURE drivetrain and a reduced-order PIPTB ballbot model. The MiaPURE drivetrain was designed to carry loads of up to 80 kg, while being comparable in size to an office chair, with a height of under 50 cm and a footprint of under 40 cm $\times$ 40 cm (approximately the width of an adult male's shoulders \cite{shah2015model}). It was designed to move alongside humans with agility similar to human walking, with a maximum speed of 2 m/s and a braking time of 2 s from a cruise speed of 1.4 m/s (the preferred walking speed of humans \cite{mcneill2002energetics}). The benchtop PIPTB version was designed to capture the unstable planar dynamics of the ballbot. It was constructed using similar hardware and served as a testbed for controller investigation and comparison.

\subsection{MiaPURE - Ballbot Drivetrain}

\begin{figure}[thpb]
  \centering
  \includegraphics[scale=0.5]{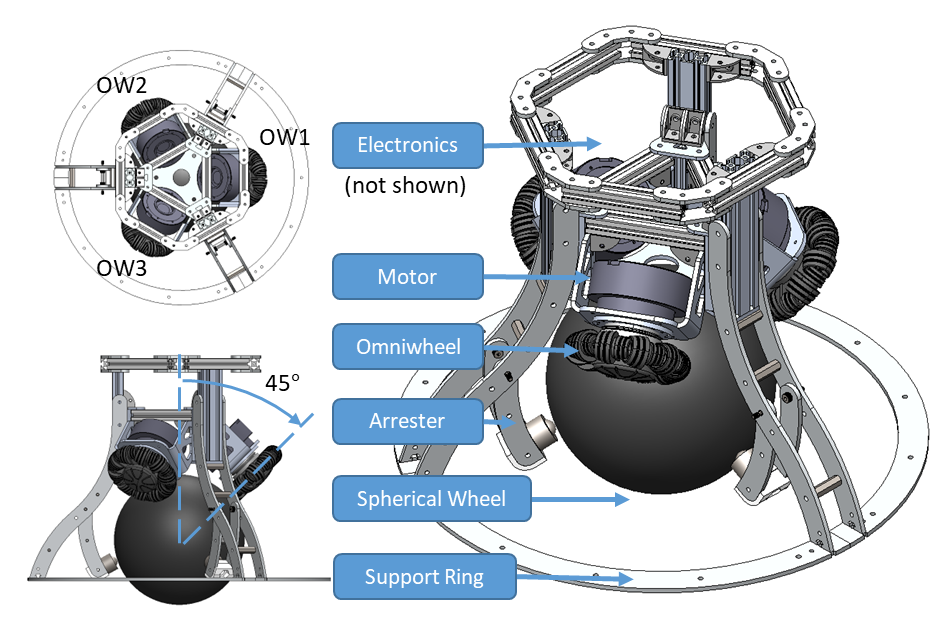}
  \caption{Mechanical design of the ballbot drivetrain for MiaPURE.}
  \label{fig0:design}
\end{figure}

The drivetrain was designed following a conventional ballbot configuration with omniwheel (OW) placement similar to BallIP \cite{kumagai2008development} and Rezero 
 \cite{leutenegger2010modeling}. Three OW-actuator pairs separated by $120^\circ$ were utilized in the design, with each OW contacting the upper surface of the spherical wheel (SW), forming a contact angle (with respect to the vertical axis) of $45^\circ$ (Fig. \ref{fig0:design}).

\subsubsection{Omniwheel}
The OWs were responsible for supporting the payload and withstanding the torque generated by the motor. For this purpose, we selected single-plate OWs with 125 mm diameter, 60 kg load capacity, and TPU-coated rollers from SeCure Inc. in China. With the proposed OW configuration, these OWs provided a theoretical static load capacity of up to 127 kg ($3\cos(\alpha)F_{LC}=127$ kg, where $F_{LC}$ is the load capacity of each OW, and $\alpha$ is the contact angle between the OW and the SW as mentioned earlier).

\subsubsection{Actuator}
We used Quasi-Direct-Drive (QDD) actuators, which are commonly used in legged robots, to provide high power density and high backdrivability \cite{seok2014design, wang2015hermes, bledt2018cheetah}. The actuators were customized using a brushless direct current (BLDC) motor and a planetary gearbox with a reduction ratio of 7.5 from T-motor Inc. in China. The resulting actuator had a diameter of 98 mm, a maximum torque of 43.2 Nm, and a no-load speed of 62 rad/s when powered by a 45 V source (without considering the efficiency of the motor driver).

\subsubsection{Spherical Wheel}
\begin{figure}[thpb]
  \centering
  \includegraphics[scale=0.5]{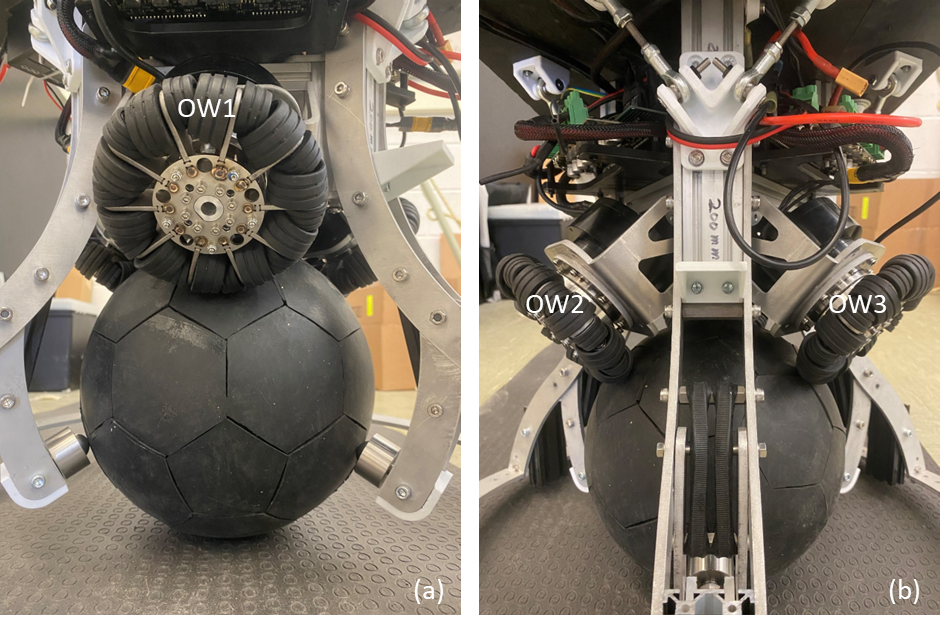}
  \caption{Physical prototype of the ballbot drivetrain viewed facing (a) OW1 and (b) OW2 and OW3.}
  \label{fig0:prototype}
\end{figure}

SW required a high load capacity, traction with the OWs, high torque transmission bandwidth, and a no-slip condition with the ground to ensure adequate yaw control authority. To meet these requirements, we fabricated our own SW using an off-the-shelf bowling ball and attached pentagon and hexagon-shaped pieces of 60A SBR rubber (with a thickness of 6.35 mm) using adhesives, as shown in Fig. \ref{fig0:prototype}. The resulting prototype is a 22.9 cm diameter SW with a weight of 3.6 kg and an estimated moment of inertia of 0.047 \si{\kilogram\milli\metre\squared}.

\subsubsection{Electrical System}
The main electronic components include a micro-controller (RoboRIO Robotics Controller, National Instrument Inc., USA), two motor drivers (ODrive V3.6, ODrive Robotics Inc., USA) to control three QDD actuators, and an inertial measurement unit (VN100, VectorNav. Inc., USA) for sensing of upper body tilting.

\subsubsection{Overview}
The final prototype of the MiaPURE drivetrain weighed 17.9 kg. The major drivetrain components including motors, OWs, and SW had a total weight of 10.4 kg, and the overall weight of the system can be further reduced with lighter chassis materials and a lighter SW. The drivetrain has a height of 45 cm (height) and a maximum footprint of 36 cm $\times$ 36 cm without the support ring (51 cm $\times$ 51 cm), which is comparable to the design target of 50 cm (height) $\times$ 40 cm $\times$ 40 cm (footprint). More details of the mechanical and electrical system of this drivetrain can be found in \cite{xiao2022ballbot}.

\subsection{PIPTB - Controller Testbed}

\begin{figure}[thpb]
  \centering
  \includegraphics[scale=0.4]{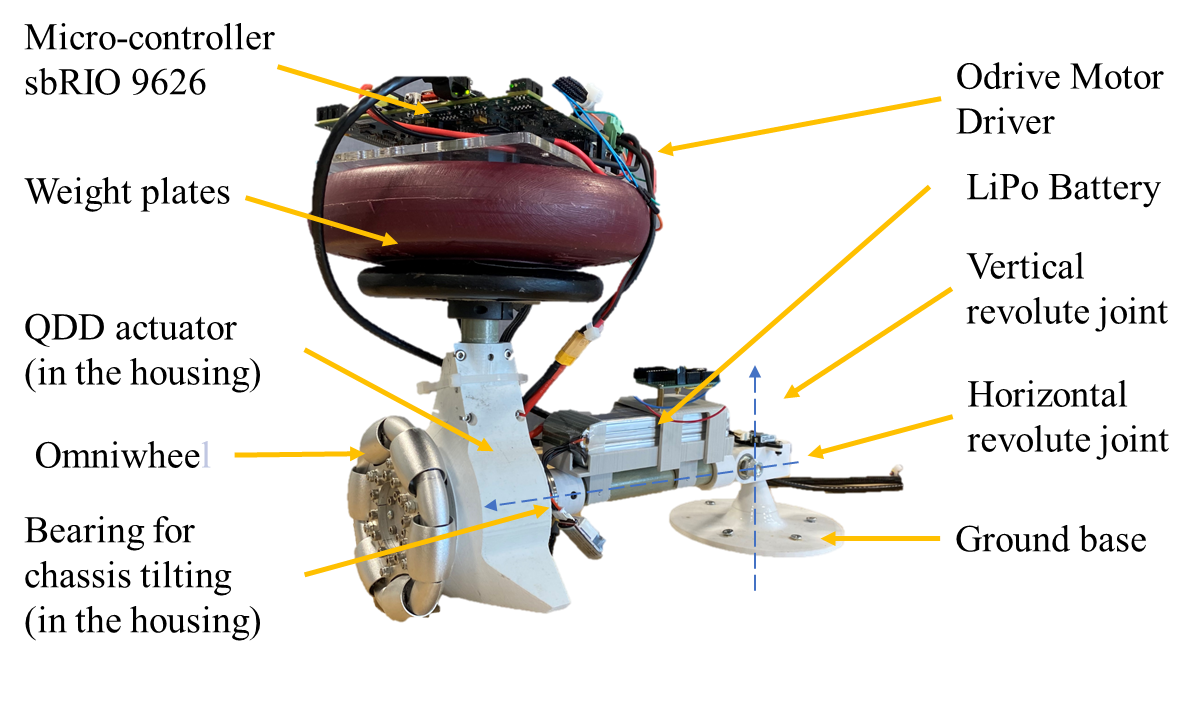}
  \caption{Physical prototype of the PIPTB while it is balancing.}
  \label{fig0:PIPTB}
\end{figure}

The PIPTB, a physical embodiment of the ballbot model, was constructed using mechanical and electrical components similar to those used in the MiaPURE drivetrain (Fig. \ref{fig0:PIPTB}). The PIPTB's design is advantageous because it can capture the unstable planar dynamics of the full-scale ballbot while lowering system complexity for experimental investigations with more controlled system parameters and environment. The PIPTB is half the size and one-fifth of the weight of the full-scale MiaPURE drivetrain (with payload). Although the testbed does not share similar static and dynamic friction properties as the full-scaled system, it can still be used to evaluate controller performances. Further details on the mechanical and electrical systems are available in \cite{xiao2022ballbot}.

\section{MODELING \& SIMULATION} \label{study0:model}

Analytical models were derived for controller development and braking performance validation. The motion of the ballbot in the 3D space was decomposed into dynamic models in three orthogonal planes. State and input trajectories for the ballbot during the braking task were further obtained using a planar model of the ballbot.

\subsection{Planar Models}

Planar models were used to describe the ballbot dynamics in the transverse, sagittal, and frontal planes, which assume negligible coupling between these three planes during spinning and translational motions (Fig. \ref{fig0:model}a). The transverse plane is perpendicular to the centerline of the upper body when fully upright, the sagittal plane is defined as the plane that intersects one of the actuators, and the frontal plane is orthogonal to the sagittal plane (Fig. \ref{fig0:model}a). 

\subsubsection{Translation Models}

\begin{figure}[thpb]
  \centering
  \includegraphics[scale=0.5]{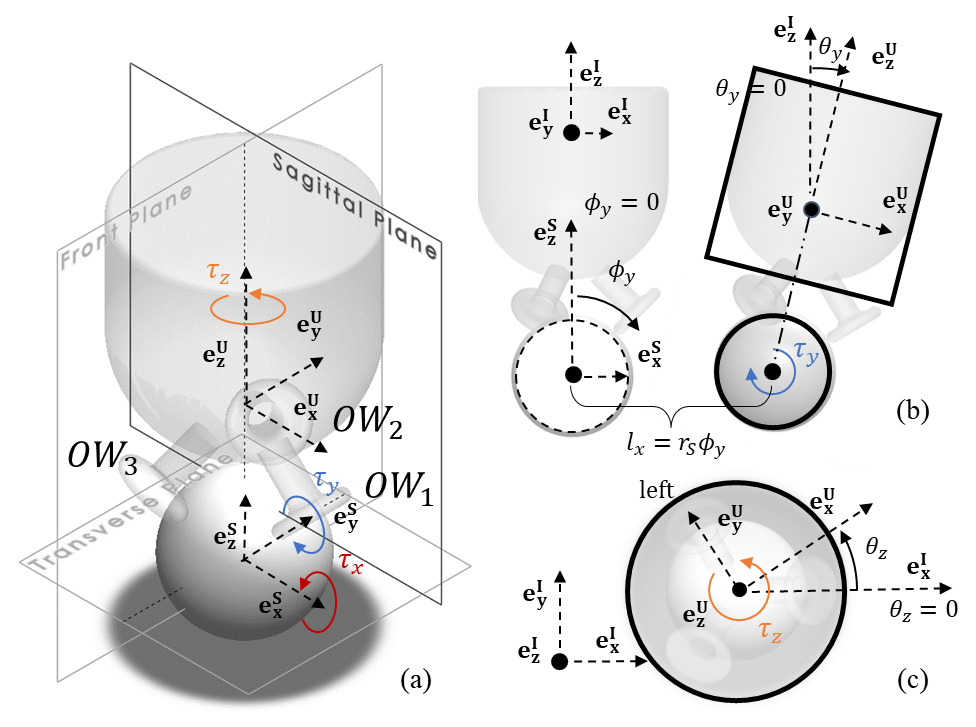}
  \caption{(a) Three individual planes defined for planar models and input torque applied to each model. (b) Translation model of the ballbot in the sagittal plane. (c) Spin model in the transverse plane, the black circle represents the upper body of the ballbot from the top view.}
  \label{fig0:model}
\end{figure}

Translational movements were decoupled into movements in the sagittal and frontal planes. The complex interactions between the OWs and SW were simplified to a torque applied to the center of the SW through a virtual revolute joint for each plane (Fig. \ref{fig0:model}b). The resultant system is a classic wheeled-inverted-pendulum (WIP) with two generalized coordinates ($\theta$ for upper body tilt angle and $\phi$ for angular position of the SW, which are each relative to the vertical axis) (Fig. \ref{fig0:model}b). The equations of motion of the WIP model in the sagittal and frontal planes were obtained using Euler-Lagrange's method \cite{spong2008robot}. The derivation of these equations is detailed in \cite{xiao2022ballbot}.

\begin{equation}
\begin{aligned}
        \boldsymbol{\ddot{q}_j} & = f_{dj}(\boldsymbol{s_j},\tau_j)\label{eqn0:trans_EOM}
\end{aligned}
\end{equation}
where $\boldsymbol{\ddot{q}_j}=[\ddot{\theta}_j, \ddot{\phi}_j]^T $, $f_{dj}(\mathbf{s_j},\tau_j)$ is the equation of motion of the WIP model, $\boldsymbol{s_j} = [\theta_j, \phi_j, \dot\theta_j, \dot\phi_j]^T$ is the state-space vector, $\tau_j$ is the torque applied to the SW, and subscript $j = y$ is for system states and input torque in the sagittal plane (i.e. $[\theta_y,\phi_y, \tau_y]$) and $j = x$ for those in the frontal plane (i.e. $[\theta_x,\phi_x, \tau_x]$).

The system dynamics of the PIPTB can be captured by the same WIP model derived in the previous section, using a generalized coordinate system of $\boldsymbol{q_P} = [\theta_P, \phi_P]^T$ where $\theta_P$ represents the chassis tilt angle relative to the pole and $\phi_p$ is the angular displacement of the wheel. Similarly, we further have $\boldsymbol{s_P} = [\theta_P, \phi_P, \dot\theta_P, \dot\phi_P]^T$ and $\tau_P$ as the state-space vector and input torque of the PIPTB system, respectively. Its equation of motion was derived as $\boldsymbol{\dot{s_P}} = f_{dP}(\boldsymbol{s_P},\tau_P)$.

\subsubsection{Spin Model}
The spinning motion of the ballbot was modeled as a single rigid body spinning around a vertical axis passing through the center of the SW in the transverse plane (Fig. \ref{fig0:model}c). Assuming no spin motion between the SW and ground \cite{lauwers2006dynamically, leutenegger2010modeling}, the system dynamics can be determined as
\begin{equation}
    I_z \ddot{q_z} + D_z(\dot{q_z})= \tau_z
    \label{eqn0:spin_dynamics}
\end{equation}
where $q_z = \theta_z$ is the yaw angle, $I_z$ is the lumped moment of inertia of the upper body and OWs, and $D_z$ represents the viscous friction torque during spinning. Detailed derivation can be found in \cite{xiao2022ballbot}. We further have the equation of motion ($f_{dz}$) of the spin model
\begin{equation}
    {\ddot{q}_z} = f_{dz}(\boldsymbol{s_z}, \tau_z)
    \label{eqn0:yaw_EOM}
\end{equation}
where $\ddot{q}_z = \ddot\theta_z$, $\boldsymbol{s_z} = [\theta_z, \dot\theta_z]^T$ is the state vector of spin model.

\subsection{Converstion to 3D Model}
For the purpose of the control system, we need to convert the state vectors and SW torques in the planar model ($\boldsymbol{s_j}, \tau_j$) into the individual OW speed and motor torque. The following conversion equations were obtained by equating the linear velocity of OWs and SW at their contacting point \cite{xiao2022ballbot}:
\begin{equation}
    \begin{aligned}
        \boldsymbol{\dot\psi} & = V_{3D}^{-1}(\dot\phi_x, \dot\phi_y, \dot\theta_x, \dot\theta_y, \dot\theta_z) \\
        \boldsymbol{u} & = T_{3D}^{-1}(\tau_x, \tau_y, \tau_z)
    \end{aligned}
    \label{eqn0:conversion}
\end{equation}
where $\boldsymbol{\dot\psi} = [\dot\psi_1, \dot\psi_2, \dot\psi_3]^T$ and $\boldsymbol{u} = [\tau_1, \tau_2, \tau_3]^T$ are respective motor speed and torque for three OW-motor pair. 

\subsection{Simulation of the Braking Task} \label{study0:dynamic simulation}

The WIP model was used to simulate the translational motion of the ballbot during the braking task: stopping from 1.4 m/s until stopped in an upright orientation within 2 s in the sagittal plane. 
The optimized trajectories for input torque ($\tau_y^*(t)$) and system states ($\boldsymbol{s_y^*}(t)$) in this task were obtained. A simple quadratic cost function $J=\int_{t_0}^{t_F} \tau_y(t)^2 dt$ was utilized to minimize the input torque during the braking task. Since these models were agnostic to a specific actuator, input torque constraints were not included and high input torque was always penalized by the objective function. 

\begin{figure}[thpb]
  \centering
  \includegraphics[scale=0.53]{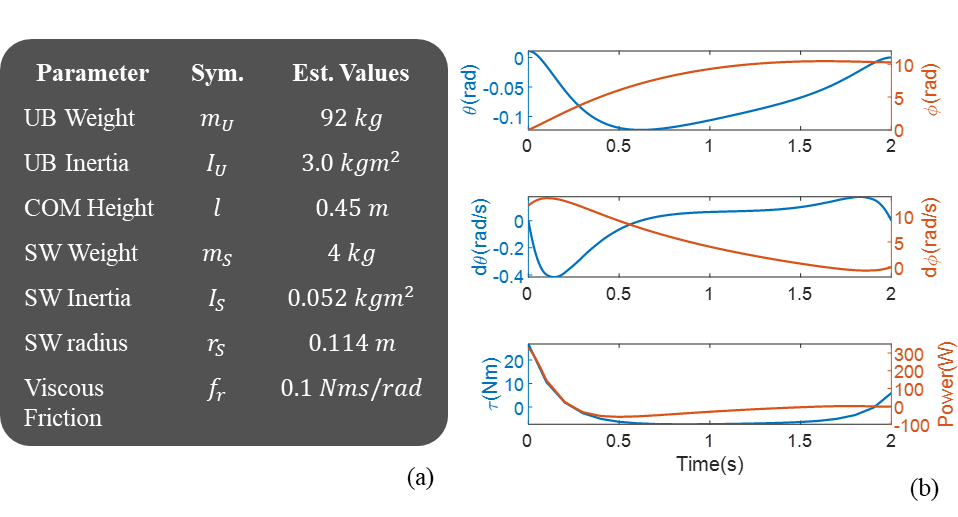}
    \caption{(a) System parameters for upper body (UB) and SW utilized for the planar ballbot model, and (b) optimized state and input trajectories of the WIP model for the braking task to stop within 2 s from 1.4 m/s.}
  \label{fig0:DEC}
\end{figure}

The formulation of the optimization is presented below:
\begin{equation}
    \begin{aligned}
\min_{\boldsymbol{s_y}[\cdot],\tau_y[\cdot]} \quad & J=\int_{t_0}^{t_F} \tau_y(t)^2 dt \\ % & \textbf{objective function}\\
\text{subject to} \quad & \boldsymbol{\ddot{q}_y}(t) = f_{dy}(\boldsymbol{s_y}(t),\tau_j(t)) \\% & \textbf{system dynamics}\\
& H(t,s_y(t),u_y(t)) \le 0  \\ % & \textbf{path constraints}\\
& G(t_0,t_F, s_y(t_0),s_y(t_F)) \le 0 \\ %& \textbf{boundary constraints}\\
\end{aligned}
\label{eqn0:optimization}
\end{equation}
where $J$ is the objective function, $\boldsymbol{s_y}$ is the state vector and $\tau_{j}$ is the input torque to the SW. The path constraint function $H(\cdot)$ includes the boundaries for system states while braking, and the boundary constraint function $G(\cdot)$ defines braking duration ($t_F-t_0$), state vector, and input at initial ($t=t_0$) and final ($t=t_F$) condition. Such an optimization problem was solved using Direct Collocation \cite{kelly2017introduction} in MATLAB, as detailed in \cite{xiao2022ballbot}. The optimized state and input trajectories for the ballbot model were generated with the following estimated system parameters (Fig. \ref{fig0:DEC}a).

Optimal solutions for system states and input torque trajectories for the braking tasks were successfully obtained (Fig. \ref{fig0:DEC}b) and were later utilized as the command state for the controller. Classic non-minimum phase behavior of the WIP dynamics can be observed from the obtained state trajectories, such that the upper body tilts backward during the braking stage, and the SW first accelerates beyond 1.4 m/s and then decelerates (Fig. \ref{fig0:DEC}b). There also exists a negative power region during the braking task, when the input torque is negative while the SW speed is positive, indicating the back-driving of the actuators.

\section{CONTROL SYSTEM DEVELOPMENT \& VALIDATION} \label{study0:control}

We aimed to develop a model-based controller that could handle the nonlinear stiction and other frictions in the ballbot drivetrain while providing intuitive tuning. However, as noted in Section \ref{study0:intro}, the open-loop torque control using model-based methods, such as LQR, used by ETH Rezero \cite{leutenegger2010modeling}, is not effective in managing high stiction in the system (Fig. \ref{fig0:controller}a). On the other hand, the cascaded PI-PD controller utilized in the CMU Ballbot \cite{lauwers2006dynamically} can break the stiction through the use of the PI controller, but it is challenging to tune (Fig. \ref{fig0:controller}b). In this study, we propose a novel controller, called cascaded LQR-PI control, that combines the strengths of both approaches. We validated the performance of this control scheme on the PIPTB before implementing it in the more complex ballbot drivetrain.

\subsection{Control System Design}

\begin{figure}[thpb]
  \centering
  \includegraphics[scale = 0.4]{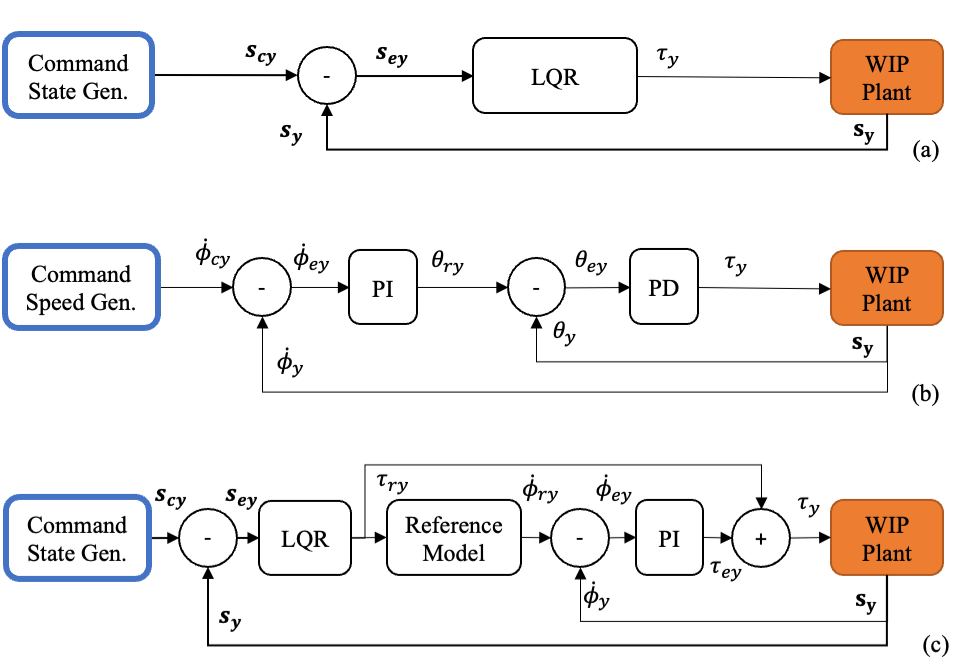}
  \caption{(a) Block diagram of an LQR torque controller for the sagittal plane WIP model, where $\mathbf{s_{cy}}$ is the command state vector, $\mathbf{s_y}$ is the measured state vector, and  $\mathbf{s_{ey}}$ is the error state vector. (b) The block diagram of a cascaded PI-PD controller, where the outer loop PI controller generates a reference tilt angle ($\theta_{ry}$) for the inner PD controller. (c) Our proposed cascaded LQR-PI controller utilizes an LQR and a reference model in the outer loop to obtain a reference speed command ($\dot{\phi}_{ry}$) for the inner loop controller, as well as a feedforward torque ($\tau_{ry}$). The subscript $y$ denotes the sagittal plane WIP model for the ballbot system, which could be replaced with $x$ for the frontal plane WIP model, or $P$ for the planar model of the WIP testbed (PIPTB) presented in the later section.}
  \label{fig0:controller}
\end{figure}

The cascaded LQR-PI controller is composed of an outer linear quadratic regulator (LQR) loop and an inner proportional-integral (PI) control loop. In the LQR loop, we utilized optimal control theory and a reference WIP model to obtain the reference SW speed ($\dot{\phi}_{r}$). The PI control loop was utilized to ensure the tracking of the reference SW speed in the WIP plant (Fig. \ref{fig0:controller}c). 

We first obtained optimal SW input torque using a linear quadratic regulator (LQR) for each planar model:  
\begin{equation}
    \tau_{rj} = \boldsymbol{k_{LQRj}}(\boldsymbol{s_{cj}}- \boldsymbol{s_j})
\end{equation}
for $j\in[x,y]$, and $\boldsymbol{k_{LQRj}}=[k_{1j}, 0,k_{2j}, k_{3j}]^T$ is the optimal LQR control gains, $\boldsymbol{s_{cj}}=[\theta_{cj},\phi_{cj},\dot{\theta}_{cj}, \dot{\phi}_{cj}]^T$ is the command state vector representing the commanded tilt angle, tilt angular rate, and SW angular speed, respectively, and $\boldsymbol{s_j} = [\theta_j, \phi_j, \dot{\theta}_j, \dot{\phi}_j]^T$ is the measured state vector for the WIP plant. The command state vector can be generated for device control with user input devices or computer-generated command state trajectories. It should be noted that we chose to not directly control the SW position since controlling the SW speed would be more intuitive for users in later applications.

In this case, we specifically have
\begin{equation}
    \tau_{rj} = k_{1j}(\theta_{cj} - \theta_j) + k_{2j} (\dot\theta_{cj} - \dot\theta_j) + k_{3j}(\dot\phi_{cj} - \dot\phi_j)
\end{equation}

The equation of motion of the reference WIP model (\ref{eqn0:trans_EOM}) was then utilized to calculate the reference SW angular acceleration ($\ddot\phi_{rj}$), which is an element of $\boldsymbol{\ddot{q}_{rj}}$, given the optimal input torque $\tau_{rj}$ and measured system state $\boldsymbol{s_j}$. We then integrated it to obtain the reference SW angular speed $\dot{\phi}_{rj}$, which is an element of $\boldsymbol{\dot{q}_{rj}}$ ($j\in[x,y]$):
\begin{equation}
    \boldsymbol{\dot{q}_{rj}} = \int{ \boldsymbol{\ddot{q}_{rj}} }dt = \int{f_{dj}(\boldsymbol{s_j},\tau_{rj})}dt
\end{equation}

The obtained reference SW speed was then utilized in an inner PI control loop to obtain a tracking torque ($\tau_{ej}$) that compensates for the speed tracking error ($\dot{\phi}_{ej}$) to ensure a zero steady-state error between the reference SW speed and the measured SW speed:
\begin{equation}
    \tau_{ej} = k_P \dot{\phi}_{ej} + k_I\int{\dot{\phi}_{ej}}dt
\end{equation}
where $\dot{\phi}_{ej} = \dot{\phi}_{rj} - \dot{\phi}_{j}$ is the tracking error of the SW speed, and $k_{Pj}$ and $k_{Ij}$ are the proportional and integral control gains for the inner PI loop. 
Finally, the total input torque for the SW is the summation of the reference input torque ($\tau_{rj}$) obtained from LQR and the tracking torque ($\tau_{ej}$) from the PI controller that compensates for unmodeled static and dynamic frictions:
\begin{equation}
    \tau_{j} = \tau_{rj} + \tau_{ej}
    \label{eqn0:control_torque}
\end{equation}

%$\dot{\phi}_{cy}, \dot{\phi}_{y}, \dot{\phi}_{ey}$ are the respective command SW speed, measured SW speed, and SW speed tracking error, $\theta_{cy}, \theta_{y},\theta_{ey}$ are the respective command tilt angle, measured tilt angle, and tilt angle tracking error.
% $\tau_{ry}$ is the reference torque input from the LQR controller, $\dot{\phi}_{ry}$ is the reference SW speed obtained using a reference WIP model, $\tau_{ey}$ is the compensation torque obtained from $\dot{\phi}_{ey}$ the inner loop PI controller to minimize the SW speed tracking error, $\tau_y = \tau_{ry} + \tau_{ey}$ is the final torque applied to the SW. 

\subsection{Controller Implementation in PIPTB}

% implementation
The LQR, the cascaded PI-PD, and the cascaded LQR-PI controllers were implemented in the PIPTB hardware. The LQR and the cascaded PI-PD controllers were directly implemented in the sbRIO controller (sbRIO9626, National Instrument Inc., USA.) at 400 Hz, and the torque command was sent to the motor driver for direct open-loop torque control. For the cascaded LQR-PI controller, the LQR outer loop was implemented in the sbRIO controller at 400 Hz. Taking advantage of the speed control capability of the motor driver, the generated reference SW speed ($\dot\phi_{rP}$) and reference motor torque ($\tau_{rP}$) were sent directly to the motor driver with the PI control inner loop running at 8 kHz. LQR control gains for the LQR and cascaded LQR controllers were generated using MATLAB (Mathworks Inc., USA), whereas the rest of the control gains were tuned manually.

\subsection{Controller Validation \& Comparison Experiment}

% Testing protocol
The performance of these three controllers was compared in a braking test. In this test, the PIPTB implemented with each controller was commanded to accelerate to 1 m/s within 2 s, hold at this speed for 2 s, and then brake within 1.4 s. A command speed was provided for the acceleration ($\dot\phi_{cP}(t) = \frac{t}{2r_W}, t\in[0,2)$) and constant speed stage ($\dot\phi_{cP}(t) = \frac{1}{r_W}, t\in [2,4)$),  where $r_W$ is the wheel radius. The optimal state trajectories for PIPTB to perform the target braking task was utilized as the command state vector during the braking stage ($\boldsymbol{s_{cP}}(t)=\boldsymbol{s_P^*}(t), t\in[4,5.4]$). Here, $\boldsymbol{s_P^*}(t)$ is the optimal state trajectory for PIPTB to brake from 1 m/s within 1.4 s, obtained using methods described in Section \ref{study0:dynamic simulation}. Three trials were repeated for each type of controller.

\subsubsection{Controller Performance Evaluation}

The resultant input torque trajectory $\tau_P(t)$ of PIPTB during the braking stage for each controller was utilized to evaluate the braking performance, using the objective function defined in the optimization problem (Section \ref{study0:dynamic simulation})
\begin{equation}
    J_P = \int_{t_2}^{t_3}{\tau_P(t)^2} dt
\end{equation}
where $t_2$, $t_3$ are the starting and ending times of the measured braking phase. We referred to $J_P$ as the braking effort, the lower value of $J_p$ is more desirable as it indicates a more efficient braking behavior. The braking effort was averaged over three trials for each controller.

\subsubsection{Results for Controller Comparison}

\begin{figure}[thpb]
  \centering
  \includegraphics[scale = 0.5]{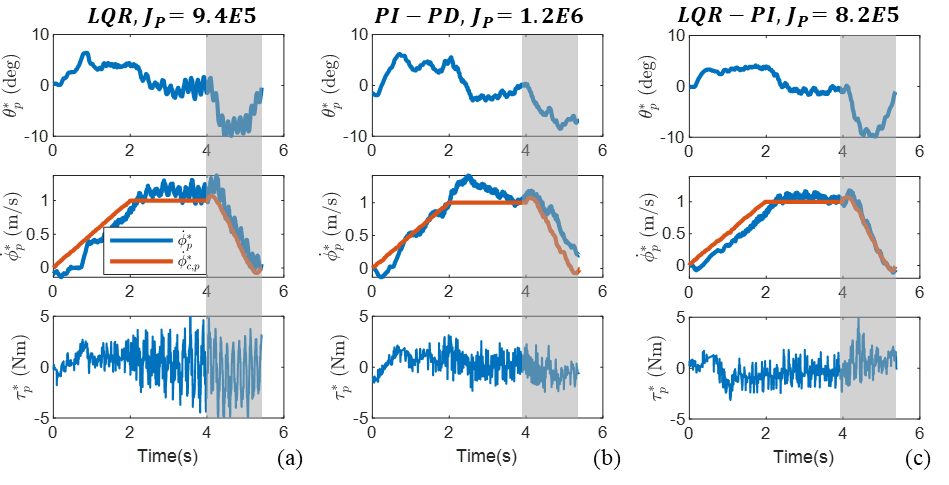}
  \caption{Resultant state ($\theta_P(t),\dot{\phi}_{P}(t)$) and input torque ($\tau_P(t)$) trajectories for PIPTB during the braking task using (a) LQR controller, (b) PI-PD controller, and (c) LQR-PI controller. The grey areas are the braking phase for these trials. $\dot{\phi}_{cP}$ is the command wheel speed of PIPTB.}
  \label{fig0:PIPTB_dcc}
\end{figure}

The resultant state and input torque trajectories for each controller in an exemplary trial are presented (Fig. \ref{fig0:PIPTB_dcc}). Among them, the PI-PD controller failed to follow the command trajectory to stop within 1.4 s from 1 m/s, hence it resulted in the highest averaged braking effort ($J_P = 1.2\pm0.3 E6$) (Fig. \ref{fig0:PIPTB_dcc}b). PIPTB with LQR and LQR-PI controllers were capable of braking within 1.4 s. LQR-PI has the lowest braking effort ($J_P = 8.2\pm0.2 E5$) due to a lower magnitude of input torque (Fig. \ref{fig0:PIPTB_dcc}c), whereas a jittery behavior was observed for PIPTB with LQR controller, making the state trajectories less smooth in the braking stage (Fig. \ref{fig0:PIPTB_dcc}a).

This simple experiment validated the feasibility of using a cascaded LQR-PI controller to control a WIP plant for quick braking. In addition, it further indicated the advantage of the cascade LQR-PI controller compared with LQR and PI-PD controller: 1) lower braking effort, 2) smoother state and input trajectories compared, and 3) better trajectory tracking capability.

\subsection{Controller Implementation in Ballbot}

\begin{figure}[thpb]
  \centering
  \includegraphics[scale = 0.5]{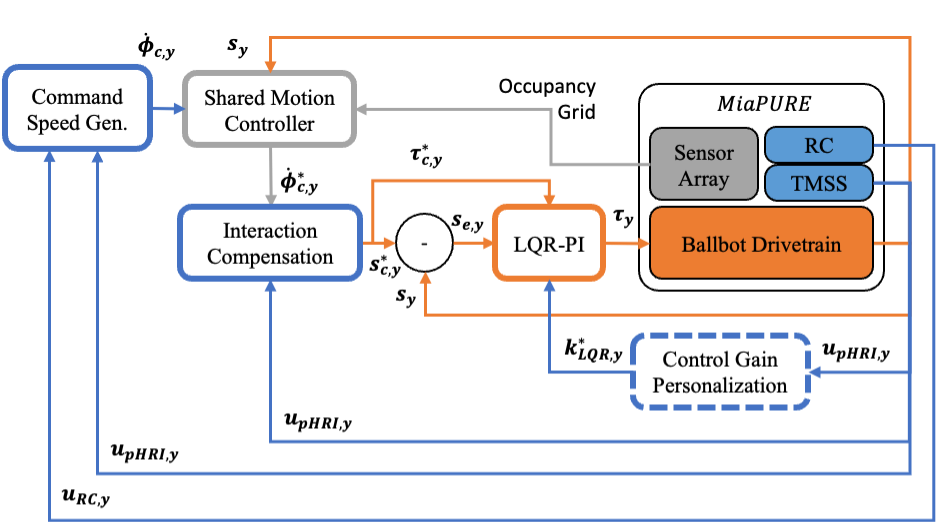}
  \caption{Implementation of the cascaded LQR-PI controller on the physical hardware of MiaPURE including the main controller running at 400 Hz and the motor driver at 8 kHz. Signals from encoders and IMU were first converted into states of planner models to obtain corresponding feedforward torque ($\boldsymbol{u_r}$) and speed command ($\boldsymbol{\dot{\phi}_r}$) of SW. They are then converted back to the feedforward torque ($\boldsymbol{\tau_r}$) and speed command ($\boldsymbol{\dot\psi_r}$) of each motor.}
  \label{fig0:PURE_control}
\end{figure}

Following the success of LQR-PI controller on PIPTB, we further implemented LQR-PI controller in the ballbot drivetrain with the same outer-inner-loop structure. The LQR outer loop was implemented in the roboRIO controller running at 400 Hz (Fig. \ref{fig0:PURE_control}). The measured upper body tilting ($\boldsymbol{\theta} = [\theta_x, \theta_y, \theta_z]^T, \boldsymbol{\dot{\theta}} = [\dot{\theta}_x, \dot{\theta}_y, \dot{\theta}_z]^T$) and OW speed ($\boldsymbol{\dot{\psi}}=[\dot{\psi}_1, \dot{\psi}_2, \dot{\psi}_3]^T$) were first coverted into state vectors all planar models $\boldsymbol{s_x} = [\theta_x, \phi_x, \dot{\theta}_x, \dot{\phi_x}]^T, \boldsymbol{s_y} = [\theta_y, \phi_y, \dot{\theta}_y, \dot{\phi_y}]^T$, and $\boldsymbol{s_z} = [\theta_z, \dot{\theta}_z]^T$ using inverse of Eqn. \ref{eqn0:conversion}.
Next, the reference input torque vector $\boldsymbol{u_{r}}=[\tau_{rx},\tau_{ry},\tau_{rz}]^T$ was obtained using implemented LQR controllers for all three planes, followed by the reference planar speed vector $\boldsymbol{\dot\phi_{r}} = [\dot\phi_{rx}, \dot\phi_{ry}, \dot\theta_{rz}]^T$ using the reference WIP models and a reference spin model. We then converted all reference torque and speed vectors in three planar models back to the reference motor torque vector $\boldsymbol{\tau_{r}}=[\tau_{r1},\tau_{r2},\tau_{r3}]^T$ and reference motor speed vector $\boldsymbol{\dot\psi_{r}} = [\dot\psi_{r1}, \dot\psi_{r2}, \dot\psi_{r3}]^T$ using Eqn. \ref{eqn0:conversion}. The resultant reference torque and reference speed for each motor were finally utilized in the PI inner loop in each motor driver running at 8 kHz to obtain the command torque to the motor using Eqn. \ref{eqn0:control_torque} (Fig. \ref{fig0:PURE_control}).

%  $\boldsymbol{s} = [ \boldsymbol{s_x}, \boldsymbol{s_y}, \boldsymbol{s_z}]^T$ and  $\boldsymbol{s_c} = [\boldsymbol{s_{cx}}, \boldsymbol{s_{cy}}, \boldsymbol{s_{cz}}]^T$ are planar state and command vectors of MiaPURE, $\boldsymbol{s_e} = [\boldsymbol{s_{ex}}, \boldsymbol{s_{ey}}, \boldsymbol{s_{ez}}]^T$ is the planar state error vector, $\boldsymbol{u_{r}}$ is the reference planar torques vector, $\mathbf{\dot\phi_{r}}$ is the reference planar speed vector, $\mathbf{\dot\psi_{r}}$ is reference motor speed vector, and $\mathbf{\tau_{r}}$ is the reference motor torque vector.}

\section{PHYSICAL ROBOT TESTING} \label{study0:exp}
% overview
The maximum speed and minimum braking time of the MiaPURE drivetrain were evaluated while carrying a payload of 60 kg, with its COM height close to that of a seated human. Only 60 kg of the payload was utilized for the safety of the device and researchers during this investigation stage. From the design target, we specified a maximum speed of 2 m/s, as well as a minimum braking time of 2 seconds when driving at 1.4 m/s. The feasibility of using the MiaPURE drivetrain for the remote control and human-riding tasks was also validated.

\subsection{Testing Protocol}
The following benchmark tests evaluated the maximum speed and minimum braking time from 1.4 m/s of the MiaPURE drivetrain. The LQR and the cascaded PI-PD controller were not formally tested due to their poor performance in the balancing task for the MiaPURE drivetrain during the pilot studies. In this case, only the cascaded LQR-PI controller was utilized to control the maneuver of the MiaPURE drivetrain for these benchmark tests. 

\begin{figure}[thpb]
  \centering
  \includegraphics[scale=0.5]{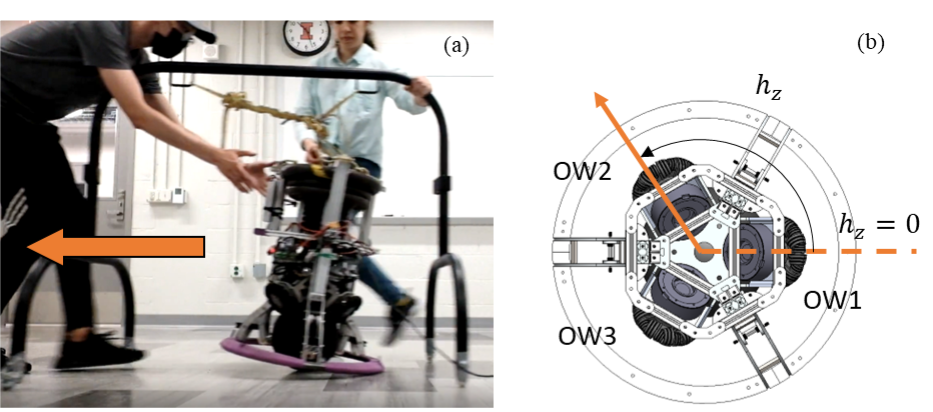}
  \caption{(a) MiaPURE drivetrain and the gantry system utilized during the benchmark experiment. A researcher pushed the gantry to follow the PURE drivetrain, while another researcher was ready to catch the robot in case of any system failure. The photo was taken during the braking of MiaPURE when translating to the left, and (b) the definition of the translation direction of the MiaPURE drivetrain, a translation direction of $0^\circ$ aligns with the OW1 axis.}
  \label{fig0:benchmark}
\end{figure}

The maximum speed was evaluated by providing a slowly ramping velocity trajectory for the balancing controller to drive the MiaPURE. The maximum speed until the system failure (losing dynamic stability) was recorded and averaged over three trials. Due to its omnidirectional maneuverability, we evaluated such system performance when the drivetrain translated towards $h_z = 0^\circ$, $90^\circ$, and $180^\circ$ (Fig. \ref{fig0:benchmark}b).

The benchmark test on the minimum braking time was only conducted if the drivetrain was capable of reaching more than 1.4 m/s stably during the maximum speed benchmark test. In this test, we commanded the robot to follow a set of trajectories composed of slow acceleration, constant speed (1.4 m/s), and braking phases. The optimal state trajectories for the braking phase were first generated with a target braking time of 5 s using methods in Section \ref{study0:dynamic simulation}. Upon successful trial completion, the robot was commanded to follow another set of state trajectories with the braking time reduced by 0.5 s, until system failure.

% results - maximum speed
\subsection{Results}
\begin{figure}[thpb]
  \centering
  \includegraphics[scale = 0.5]{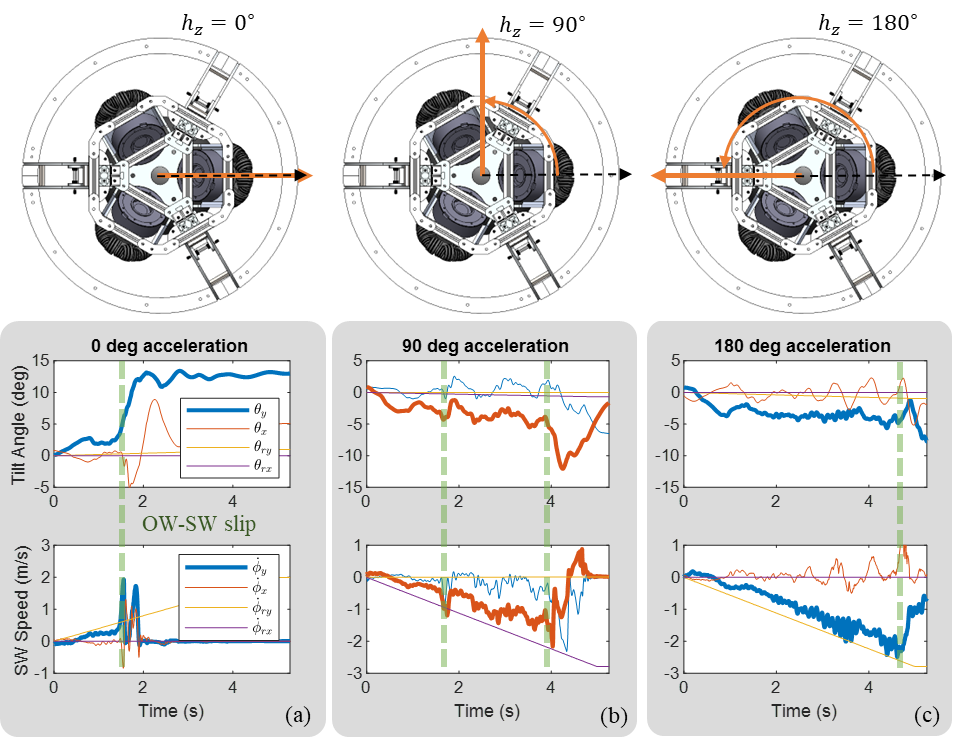}
  \caption{Measured system responses of tracking an acceleration profile when translating in (a) $h_z = 0^\circ$, (b) $h_z = 90^\circ$, and (c) $h_z = 180^\circ$. Visualizations of the translation direction relative to the MiaPURE drivetrain are presented in the figures above the plot. The instant before the system failure or slip was marked with a dashed green line. Among these translational directions, only the last one ($h_z = 180^\circ$) showed a promising result of reaching beyond 2.0 m/s before failure.}
  \label{fig0:acc_study0}
\end{figure}

\begin{figure}[thpb]
  \centering
  \includegraphics[scale = 0.5]{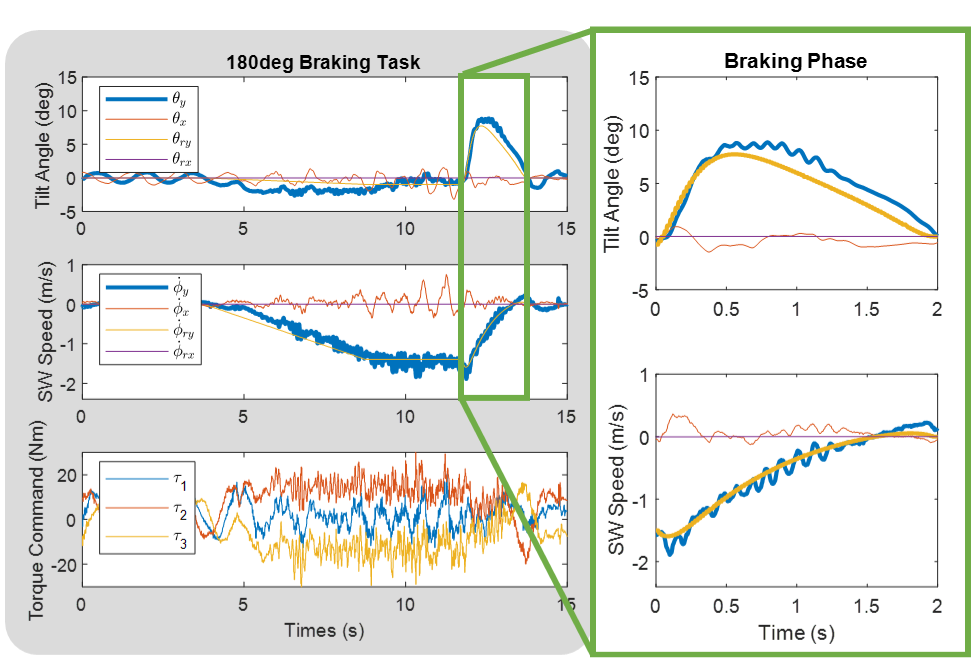}
  \caption{Measured system response of the PURE drivetrain tracking a braking profile when driven in a $180^\circ$ heading angle. The device successfully decelerated from 1.4 m/s within 2 s.}
  \label{fig0:dcc_study0}
\end{figure}

The MiaPURE drivetrain demonstrated distinct maximum speeds for different translation directions. Translating in $h_z = 0^\circ$ resulted in the lowest maximum speed of less than 0.6 m/s (Fig. \ref{fig0:acc_study0}a), while moving in $h_z = 180^\circ$ resulted in the highest maximum speed of 2.3 m/s (Fig. \ref{fig0:acc_study0}c). The braking experiment was only performed with the PURE drivetrain translating in $h_z = 180^\circ$. The robot was capable of decelerating with a minimal braking time of 2 s (Fig. \ref{fig0:dcc_study0}).

\subsection{Demonstrations}

\begin{figure}[thpb]
  \centering
  \includegraphics[scale = 0.7]{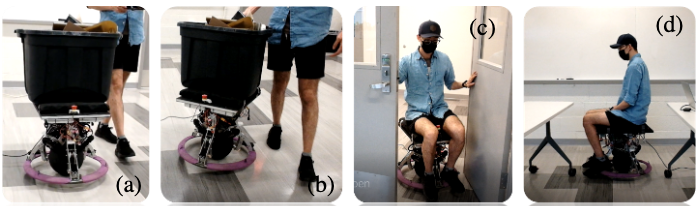}
  \caption{Demonstration of MiaPURE drivetrain for (a,b) payload carrying and (c,d) human riding.}
  \label{fig0:demo_study0}
\end{figure}

We further assessed the feasibility of using the MiaPURE drivetrain as a payload-carrying robot and a human-riding device to provide hands-free assistive mobility. As the payload robot, MiaPURE can be controlled either through a remote control device (RC) or physical human-robot interactions (pHRI) by gently pushing on the payload to generate an omnidirectional maneuver (Fig. \ref{fig0:demo_study0}a, b). As the riding device, the rider can utilize torso leaning to control its omnidirectional maneuver (sliding through a narrow space), while utilizing hands for more important tasks such as door opening (Fig. \ref{fig0:demo_study0}c, d). The demo video can be found via this link: \url{https://www.youtube.com/watch?v=H3Wc7nfBxZ8}.

\section{DISCUSSION}\label{study0:discussion}
%\subsection{on LQR-PI Controller}
The LQR-PI controller took advantage of physical hardware in our platforms. The LQR outer loop (400Hz) handles the slower whole system dynamics and performs a single-step forward simulation to generate feedforward torque and speed command for motors. The PI inner loop (8kHz) mainly deals with faster and more complicated motor dynamics and unknown frictions in the OW-SW interaction. More, IMU used for LQR outer loop also has a lower updating rate than encoders used for PI inner loops. In this case, such an outer-inner loop structure help to handle both balancing and stiction-compensation problem, outperforming LQR and PI-PD controllers in both the PIPTB and ballbot drivetrain.
%\subsection{on Benchmark Test}

It was surprising to observe that the speed performance of the ballbot depended on the translation direction (Fig. \ref{fig0:acc_study0}). Indeed, one would tend to initially think that the ballbot should have similar performance in all directions due to its spherical wheel and neglect that the proper actuation of the spherical wheel requires a friction cone constraint \cite{erdmann1994representation} for each OW-SW contact point. For example, translating in $0^\circ$ and $180^\circ$ have the same magnitude of traction force on two driving OW2 and OW3, while OW1 is mainly idling. However, the normal forces on two driving OWs have smaller magnitudes when translating in $0^\circ$ due to the chassis leaning towards OW1. Such an effect reduced friction cones on OW2 and OW3, causing slip in OW-SW contacts and leading to loss of balancing.
%\subsection{on Human-Riding}

We further demonstrated the feasibility of using the MiaPURE robot for human riding and navigation in a constrained space. The researcher was capable of controlling the translational movement of MiaPURE in a completely hands-free mode via torso leaning, similar to riding a Segway device. However, due to the high risk of failure when translating in $0^\circ$ (seat back direction), it is not feasible to perform a quick translation towards the back when riding this drivetrain without slipping between OWs and SW. The contact stability between OWs and SW needs to be improved to allow for agile translation in all directions.
%\subsection{Future Work}

The limitations of the current device indicated several important directions for future work. The system failure due to the slip between OWs and SW of the device needs to be mitigated to ensure that all translational directions are safe. We need to investigate the fundamental drivetrain mechanism to understand how ballbot drivetrain design variables affect contact stability between OWs and SW during the safety-critical tasks and iterate drivetrain design to ensure system safety.

\section{CONCLUSION}\label{study0:conclusion} 

In this paper, we presented our attempt to build the mechanical and control system of a high load capacity, minimal footprint mobile robot using the technology of a ballbot. To compensate for unmodeled friction in the torque transmission system, we developed a cascaded LQR-PI controller to help with overcoming the stiction and dynamic friction in the prototype, in addition to allowing for intuitive gain tuning. The controller was first validated in a planar inverted pendulum testbed and then implemented on a full-sized physical prototype. The prototype was capable of achieving a maximum speed of 2.3 m/s and braking within 2.0 seconds from 1.4 m/s while carrying a static payload of 60 kg. In addition, we further demonstrated the feasibility of a human riding the MiaPURE drivetrain using torso leaning through demonstrations of manipulation (door opening) and locomotion tasks (sliding through a narrow space). These results highlight the potential of building a ballbot drivetrain, with the proposed design, fabrication, and control methodology, as a universal mobility platform for various tasks that require handling heavyweight with high COM while navigating in a constrained environment.

\section*{Acknowledgment}

The authors thank Coach Adam Bleakney, Doctor Jeannette Elliot, Professor Deana McDonagh, Professor William Norris, Doctor Patricia Malik, graduate students Yu Chen and Seung-Yun (Leo) Song, and undergraduate student Zheyu Zhou for their help and support with concept generation, physical hardware development, and system testing.

%\begin{thebibliography}{00}
\bibliographystyle{./IEEEtran} % use IEEEtran.bst style
\bibliography{./IEEEabrv,./reference}
%\end{thebibliography}

\end{document}